\documentclass[journal]{IEEEtran}
\ifCLASSINFOpdf
\else
\fi
\usepackage{xspace}
\usepackage{amssymb}
\def\etc{\emph{etc}\@\xspace}

\def\etal{\emph{et al}\@\xspace}
\def\ie{\emph{ie}\@\xspace}

\usepackage{graphicx}
\usepackage{etoolbox}
\makeatletter
\patchcmd{\@makecaption}
  {\scshape}
  {}
  {}
  {}
\makeatletter
\patchcmd{\@makecaption}
  {\\}
  {.\ }
  {}
  {}
\makeatother
\def\tablename{Table}
\usepackage{color,xcolor}

\begin{document}
%
\title{Instance and Pair-Aware Dynamic Networks for Re-Identification}
\graphicspath{{image/}}
%
%
%

\author{Bingliang Jiao*, Xin Tan*, Jinghao Zhou*, Lu Yang, Yunlong Wang, Peng Wang$\dagger$\\School of Computer Science, Northwestern Polytechnical University, Xi’an, China.
\thanks{The first three authors contribute equal to this work. }}

%
%

\markboth{Journal of \LaTeX\ Class Files,~Vol.~14, No.~8, August~2015}%
{Shell \MakeLowercase{\textit{et al.}}: Bare Demo of IEEEtran.cls for IEEE Journals}
%



\maketitle

\begin{abstract}

Re-identification (ReID) is to identify the same instance across different cameras. Existing ReID methods mostly utilize alignment-based or attention-based strategies to generate effective feature representations. However, most of these methods only extract general feature by employing single input image itself, overlooking the exploration of relevance between comparing images. To 
fill this gap, we propose a novel end-to-end trainable dynamic convolution framework named Instance and Pair-Aware Dynamic Networks in this paper. The proposed model is composed of three main branches where a self-guided dynamic branch is constructed to strengthen instance-specific features, 
focusing on every single image. Furthermore, we also design a mutual-guided dynamic branch to generate pair-aware features for each pair of images to be compared. Extensive experiments are conducted in order to verify the effectiveness of our proposed algorithm. We evaluate our algorithm in several mainstream person and vehicle ReID datasets including CUHK03, DukeMTMC-reID, Market-1501, VeRi776 and VehicleID. In some datasets our algorithm outperforms state-of-the-art methods and in others, our algorithm achieves a comparable performance. 

\end{abstract}	


%
\IEEEpeerreviewmaketitle

\section{introduction}\label{sec_Int}
\IEEEPARstart{R}{e-identification} (ReID) is a fundamental and challenging task in computer vision, aiming at matching a specific instance in distributed and non-overlapping camera views. Given a query image and a large set of gallery, ReID represents each image as a feature embedding and ranks the gallery in the light of the feature similarity between the query and images in the gallery, thus obtaining the closest match.\par
Due to its application in fields like intelligent surveillance and tracking, ReID has attracted explosive attention in the community and made great progress.
Nevertheless, there are several challenges that still exist in the practical unconstrained scenarios.
Those challenges may arise from view misalignment, background perturbance, 
view-angle changes, pose variations and noisy labels,~\etc.
\begin{figure}[t]
	\begin{center}
		\includegraphics[width=1.0\linewidth]{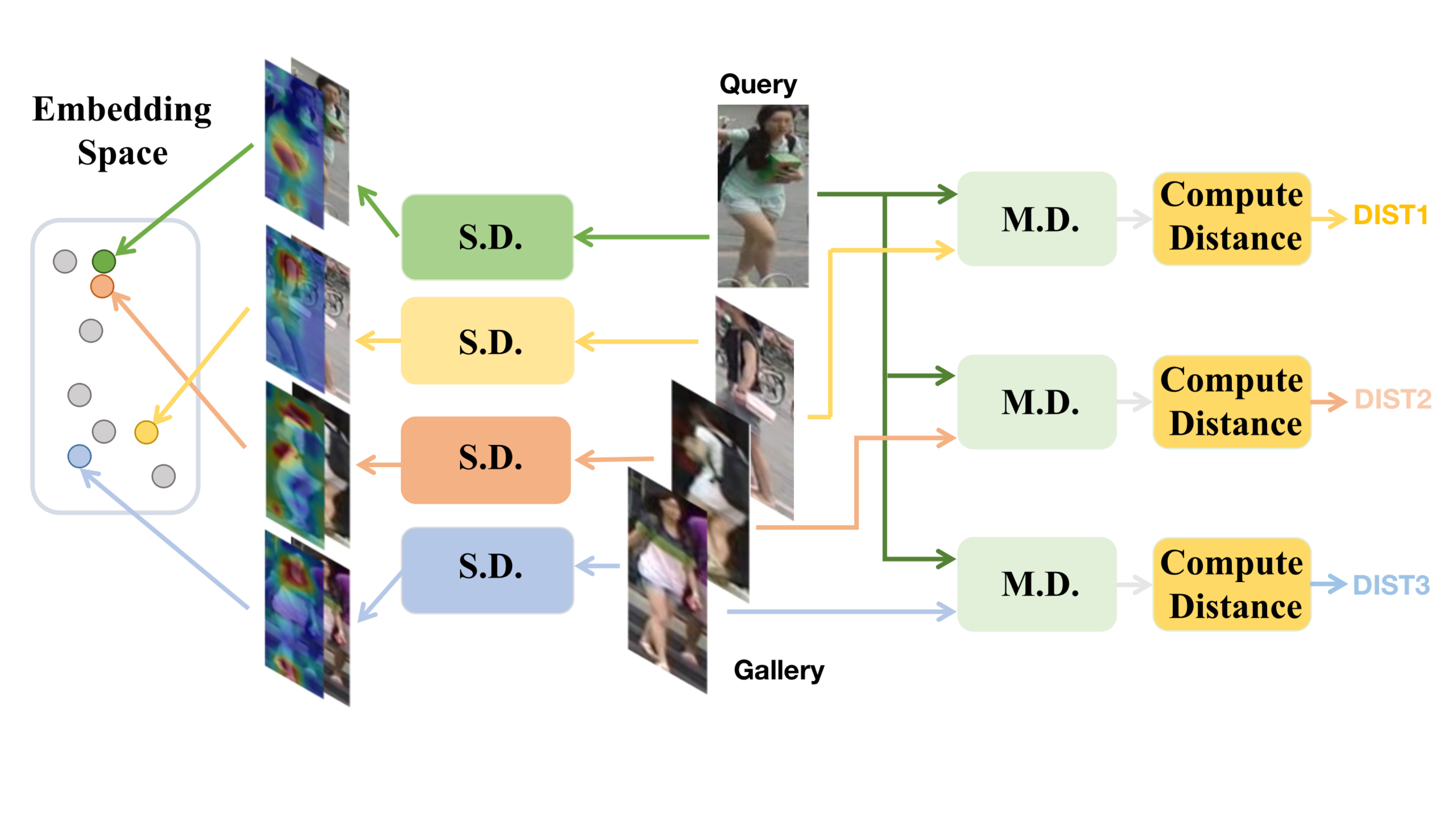}
	\end{center}
	\vspace{-5mm}
	\caption{
	The sketch above displays the mechanism of our primary model branches, namely Self-Guided Dynamic branch (S.D.) and Mutual-Guided Dynamic branch (M.D.). Each input image will obtain its instance-level specific and discriminative features through S.D. and then be projected to a shared embedding space where the same IDs (such as the query and the middle one in the gallery ) are more likely to be clustered for the sake of the powerful representation competence of S.D. module. Simultaneously, each image in the gallery will be fed into the M.D. with the query respectively to exploit their pairwise semantic relevance and measure the distance for matching procedure.
	}
	\label{fig:1}
\end{figure}
Tremendous efforts have been devoted to overcoming the mentioned difficulties. Some algorithms~\cite{sun2018beyond,guo2019beyond} fuse the body part information to construct the final features for mitigating the impact of misalignment and occlusion. Attention-based methods
~\cite{chen2019abd,zhang2020relation,si2018dual} have also been widely used to strengthen the noteworthy areas and suppress the interference via generating highly context-related responding masks.
However, all those 
methods represent each image in a fixed mould and can hardly 
flexibly learn the instance-level specific feature which is crucial for instance identifying.
Moreover, the mutual correlation between the 
instances to be identified is given few consideration during the process of extracting feature in the existing researches.

To tackle these issues above, in this paper we propose a novel three-branch Instance and Pair-Aware Dynamic Network (IPAD) as shown in Figure~\ref{fig:1}. For ReID algorithm, it is significant to obtain discriminative and instance-specific feature for each input. Considering the lack of flexibility and specificity in regular convolution operations, we apply the dynamic convolution blocks to further facilitate feature representation, instead of merely resorting to static filters. Through those dynamic layers, we can augment representation capability of the model with increasing slight
parameters. On the basis of this theory, we creatively design the self-guided dynamic branch and the mutual-guided dynamic branch.
The self-guided dynamic branch aims to extract specific and discriminative features 
via customized kernels for each input image.
And the purpose of mutual-guided dynamic branch is to capture the subtle but mutually discriminative visual cues between pairwise images.
In such network architecture, different guidance instances it employs provide complementary perspectives to achieve self-adaptivity and mutual-adaptivity, thus gaining sightly performance improvement.

In summary, we have made two major contributions:  
\begin{itemize}
\item We propose a novel and powerful three-branch network IPAD, incorporating a self-guided dynamic branch, a global branch and a mutual-guided dynamic branch. Such framework can mutually provide complementary information and assist to enhance the robustness of our model. 
Moreover, the applicability and compatibility of IPAD have been validated through carefully designed experiments.
\item Rich ablation study has been conducted to prove the effectiveness of our novel and delicate model branches. And extensive experiments demonstrate the superiority of
the proposed IPAD over a wide range of ReID models on several mainstream benchmarks.
\end{itemize}
\par

The rest of the paper is organized as follows: Section~\ref{sec_RW} reviews the related works of our paper. In Section~\ref{sec_method} we elaborate the proposed ReID method. The experimental results are reported in Section~\ref{sec_expr}. And we conclude the paper in Section~\ref{sec_conc}.

\hfill XXX
\hfill XXXX XX, 20XX

\section{Related Work}\label{sec_RW}
  In this section, we review some recent researches referring the dynamic convolution and Re-identification.
\subsection{Dynamic Convolution}
In recent researches, more and more attention has been paid to the use of dynamic mechanism in various computer vision tasks.

\begin{figure*}[t]
	\begin{center}
		\includegraphics[width=1.0\linewidth]{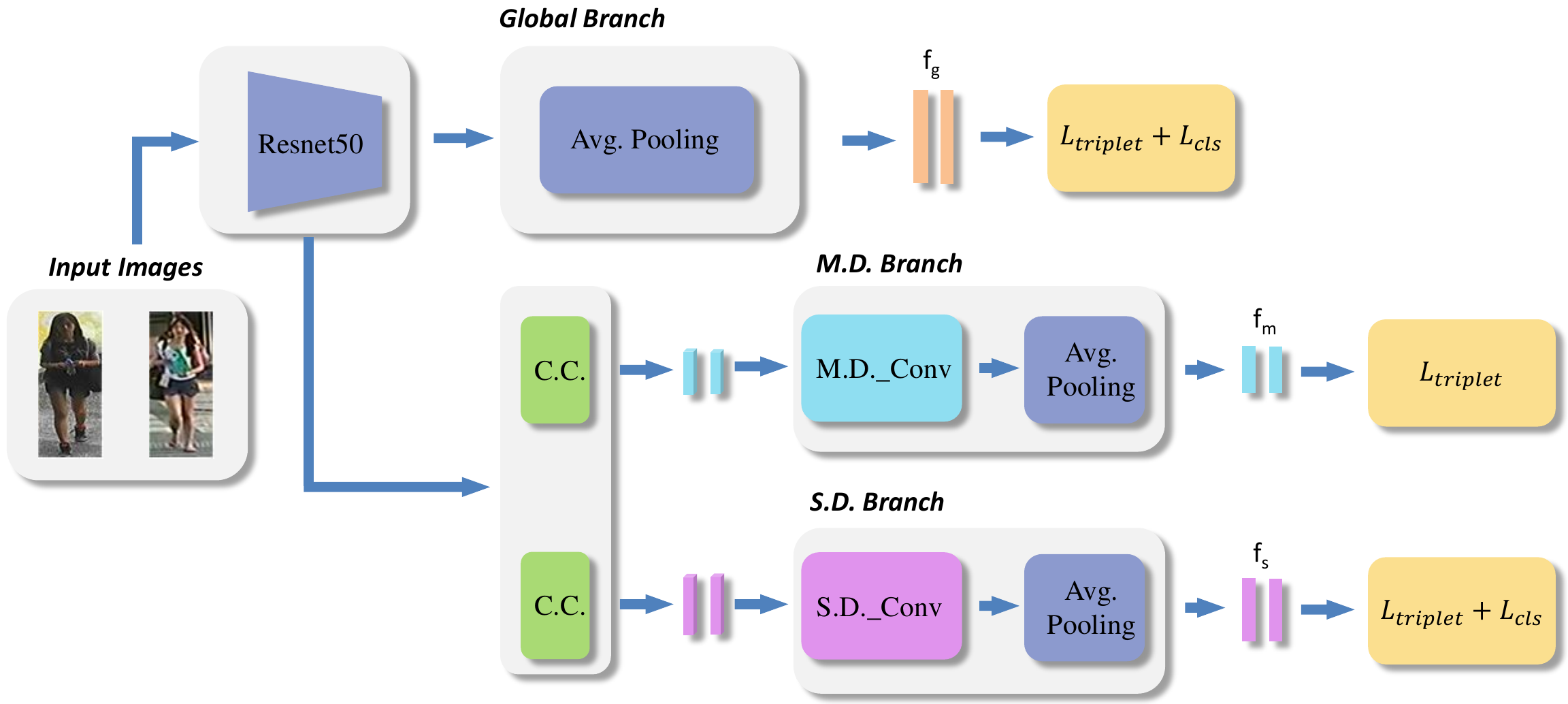}
	\end{center}
	\vspace{0.05cm}
	\caption{
		The overall structure of our proposed algorithm. The ``S.D.", ``M.D.", ``C.C.", ``$L_{triplet}$" and ``$L_{cls}$" represent Self-Guided Dynamic, Mutual-Guided Dynamic, Channel Compression Module, Triplet Loss and ID Loss respectively.}
	\label{fig:structure}
\end{figure*}

Yang ~\etal~\cite{yang2019condconv} proposed to replace the traditional convolution by the conditionally parameterized convolution, in the purpose of improving the performance and inference cost trade-off of some existing neural networks. 
Chen ~\etal~\cite{chen2020dynamic} pointed that aggregating those kernels in a non-linear way via attention makes the convolution more powerful in terms of representation competence. 
Zhang ~\etal~\cite{zhang2020dynet} introduced adaptive generated convolution kernels according to the images contents in order to reduce the redundancy among previous static filters and performs outstandingly. 

Above all, it is reasonable to infer that dynamic mechanism does contribute to produce more content-adaptive filters and  more instance specific features.
It can make the model more flexible and adaptive to the diverse inputs.
In our algorithm, we not only apply the regular self-guided dynamic convolution module, but further dig into the mutual-adaptivity between images and therefore construct a mutual-guided dynamic branch.
\subsection{Re-Identification}
Re-Identification is a challenging task to associate individual identities across different time and locations. Nowadays, ReID has embraced abundant attention from researches in computer vision.\par
\textbf{Person \ Re-identification. }Many algorithms make use of both global features and local features simultaneously to exploit their respective advantages.
Wang~\etal~\cite{wang2018learning} carefully designed a multiple granularity network, which
consists of one branch for global features and two branches
for local feature representations. 
Several algorithms devote to overcoming impact from inaccurate person detection via alignment-based methods.
PCB ~\cite{sun2018beyond} is a method based on horizontal splits to match the part features in the purpose of suppressing the occlusion and tackling the misalignment. Besides the algorithms mentioned above, a serials of attention-based algorithms ~\cite{zhang2020relation,si2018dual,chen2019mixed,yang2019towards} have been employed in ReID and made competitive performance through augmenting the most salient feature area through correlation analysis.
Zhang ~\etal~\cite{zhang2020relation} introduced an RGA model which learns the correlation/affinity from all the feature positions and feature itself.
In addition, Si~\etal~\cite{si2018dual} proposed a DuATM network which conducts feature comparison with dual attention mechanism.
There are many other ReID works~\cite{suh2018part,kalayeh2018human,wang2018mancs,wang2018learning,miao2019pose,liao2020interpretable}, as well.

\textbf{Vehicle\ Re-identification. }Vehicle ReID has prevailed since the past few years.
Khorramshahi~\etal~\cite{khorramshahi2019dual} utilized 20 vehicle keypoints to generate attention maps by categorizing keypoints into four groups, yet the keypoint annotation is hard to acquire in real-world scenarios. 
Wang ~\etal~\cite{wang2019vehicle} proposed an attention based framework which is able to capture the discriminative parts of vehicle instance with via additional annotation.\par
Although those mentioned approaches did make effects on ReID tasks, they mostly learn the representation in a fixed module.
Thus the mapping relationship in their network is not strong and diverse enough to represent different images in the way according to the contents of inputs, making the algorithm less flexible and the learned feature less distinguishable. Even for the pair-aware attention-based module such as the DuATM, it only adopts the weighted pooling mechanism for feature augmentation whose transformation function is not strong enough to produce instance-specific feature representations. 
Several other algorithms based on pair-aware matching have also been presented, such as Contrastive CNN~\cite{han2018face} designed by Han~\etal, which explores the difference between local grid of comparing face images to perform contrastive convolution. However, the misalignment of instance to be identified makes it not applicable to ReID task.

As for our IPAD, the dynamic module applied in our algorithm generates customized filters for each input image which contribute to much more complex mapping relationship. This module makes for the constuction of more diverse and specific features and benefit the matching process.

In addition, 
we constitute a cross-referring mechanism to exploit mutually discriminative cues via introducing mutual-guidance in the feature extraction phase. Combining the cross-referring mechansim and dynamic module together, our algorithm presents strong capability and robustness.

\section{Approach}\label{sec_method}
In this section we elaborate on the detail of the proposed Instance and Pair-Aware Dynamic Networks (IPAD).
Our algorithm is detailed in the following lines.

\begin{figure}[t]
	\begin{center}
		\includegraphics[width=1.0\linewidth]{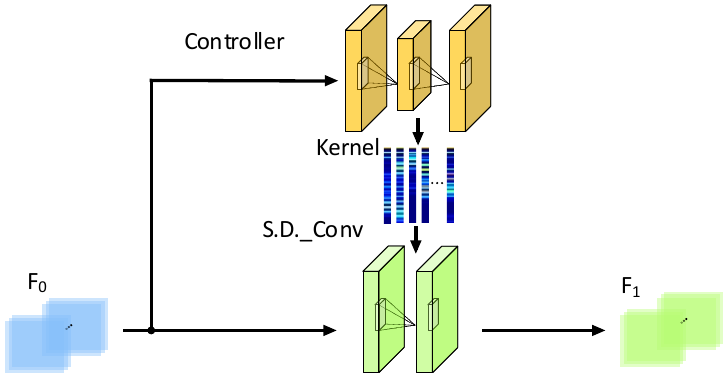}
	\end{center}
	\vspace{0.05mm}
	\caption{
		The sketch of our Self-Guided Dynamic Convolution Module. The controller employed in our model is a shallow network. The ``S.D.\_Conv" represents Self-Guided Dynamic Convolution layer while the ``$F_0$/$F_1$" represent the feature before/after the dynamic convolution layer respectively.}
	\label{fig:dynamic_conv}
\end{figure}

\subsection{Overall Framework}
In this paper, we seek to explore an effective algorithm which is able to capture the discriminative person features, so as to enhance the capability of our model. 
Besides, we argue that the discriminative feature of each person image is compose with two-folds, \ie the salient areas of input image itself and the distinguishable areas with other compared image.
In order to adequately capture these discriminative features, we design a Instance and Pair-Aware Dynamic Network (IPAD) consists of three branches, namely global branch, self-guided dynamic branch and mutual-guided dynamic branch.

In our algorithm, the global branch is a vanilla convolution module which adopts the Resnet50~\cite{he2016deep} pretrained in ImageNet~\cite{deng2009imagenet} as the backbone model.
The extracted naive convolutional features (global features) from this branch will then be send to the other two dynamic branches for further refinement.
In the self-guided dynamic branch, we design a dynamic convolution module as illustrated in Figure~\ref{fig:dynamic_conv}.
This branch could capture salient areas of each input person image by a carefully designed self-guided dynamic module.
In employed dynamic module, a controller module is constructed to transform global features to a series of customize dynamic kernels.
Thereafter, we adopt these generated kernels which preserve the identity information to perform convolution on global features, 
so as to capture the 

Beyond the self-guided dynamic branch, a mutual-guided dynamic branch has been established to obtain pair-aware instance-specific features for each pair of input images.
In this branch, each pair of input images mutually provide dynamic kernels for further feature refinement. In this cross-referring mechanism, images belonging to the same ID tends to be represented more closely in the embedding space and those not would show significant difference thanks to the customized kernels.

Extensive ablation studies we designed demonstrate the mutual-guided dynamic branch and the self-guided dynamic branch could mutually provide complementary information and further reinforce the capability of our model.


\subsection{Global Branch}

The global branch is designed to extract holistic-level feature by adopting a simple global average pooling over the convolution maps generated by the backbone model. Besides, a batch normalization layer is inserted after the pooling layer in order to ensure convergence.
Meanwhile, we utilize two widely used losses to train the ReID module, namely Triplet Loss and Cross Entropy Loss.
The loss function of this branch can be formulated as,
\begin{equation}
L_{global} = L_{triplet} + L_{cls}
\end{equation}
Noting that in all the experiments mentioned in the manuscript, the Label Smooth Algorithm~\cite{szegedy2016rethinking} and the Hard Sample Strategy~\cite{hermans2017in} have been adopted to enhance the performance of our model.

\subsection{Self-guided Dynamic Branch}
In this subsection, we expand on the detail of the self-guided dynamic branch in our algorithm. 
Different from the regular convolution layer whose kernels are static and shared among all input images, our self-guided dynamic convolution module generates customized kernels for each input image.
The sketch of our self-guided dynamic branch is shown in Figure~\ref{fig:dynamic_conv}. It can be seen from the Figure~\ref{fig:dynamic_conv}, the self-guided dynamic branch is composed of two components, termed controller module and self-guided dynamic convolution layer. 
Firstly, the controller module will be described in detail.
In fact, the controller can be regarded as a nonlinear transformation module which converts the holistic-level feature to dynamic convolution kernels.
Therefore, we adopt a shallow network as the controller which can be formulated as following,
\begin{equation}
\pi (X)=\sigma (WX+\beta)
\end{equation}\label{fun:MLP}
\noindent where the $\pi()$ represents the shallow network module, the $\sigma$ represents ReLU activation, the $\beta$ and $W$ are 
trainable parameters representing the biases and a three-step transformation function mapping $\mathbb{R}^{C_o} \mapsto  \mathbb{R}^{C_o/d} \mapsto \mathbb{R}^{C_o/d} \mapsto  \mathbb{R}^{C_t}$ with $C_o$, $d$ and $C_t$ being the original feature dimension, squeezing rate and output feature dimension. 
Thereafter, we adopt the F2 Normalization over the generated kernels to ensure convergence.

\begin{figure}[t]
	\begin{center}
		\includegraphics[width=1.0\linewidth]{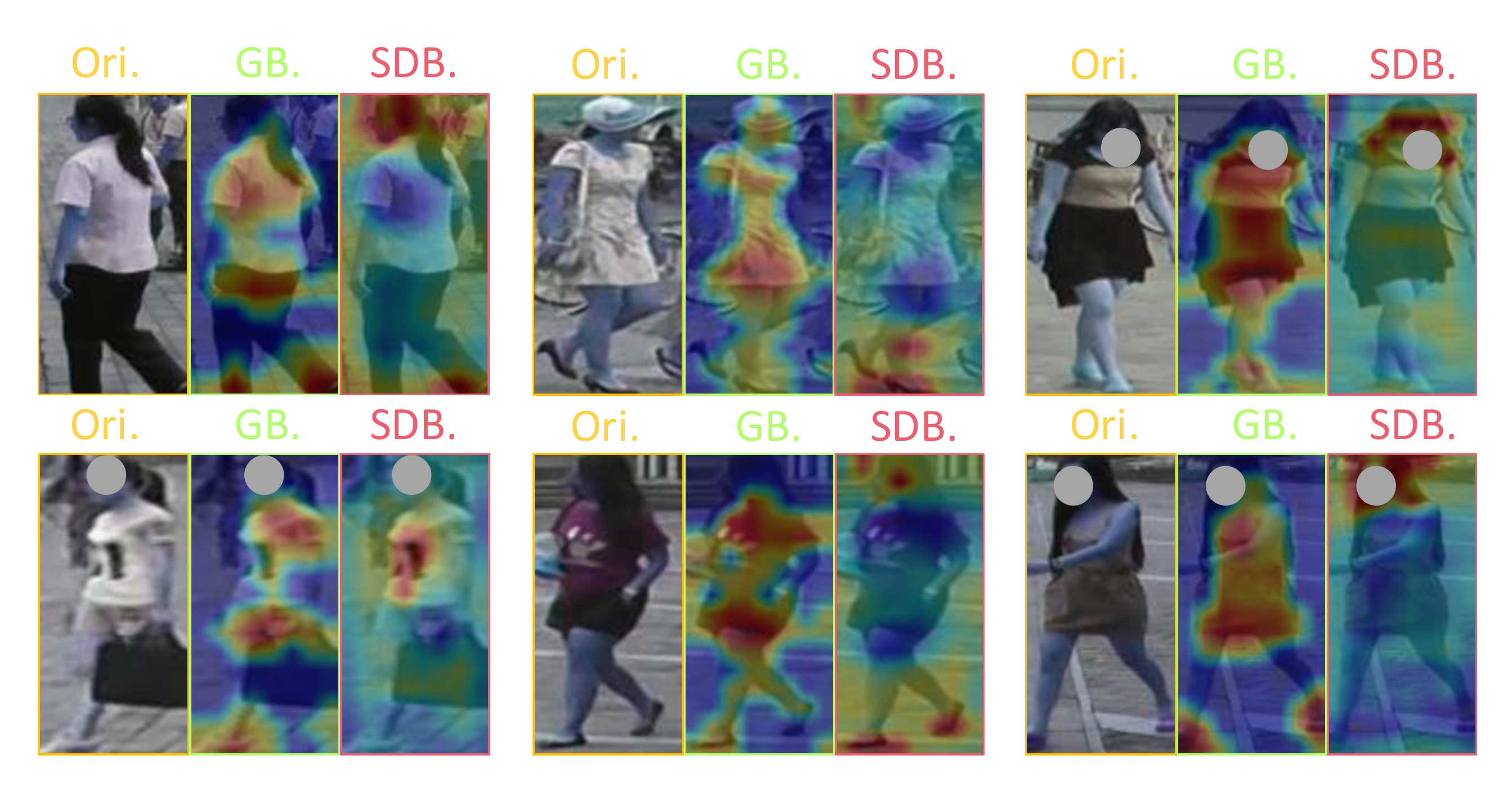}
	\end{center}
	\vspace{-5mm}
	\caption{
		The heatmaps generated by our proposed model. The ``Ori.", ``GB.", ``SDB." represent the Original Image, Heatmap from Global Branch and Self-Guided Dynamic Branch respectively.}
	\label{fig:heatmap}
\end{figure}

In parallel with the controller module, the detail of the dynamic convolution layer will be described.
During the feature extraction phase, we firstly compress the dimension of holistic-level feature from $2048$ to $512$ through a shallow network to mitigate the computational consumption.
In our algorithm, three layers of convolution inserted with ReLU and batch normalization have been employed as the channel compression module.
Thereafter, we construct a $1\times1$ dynamic convolution layer whose input dimension and output dimension are both $512$.
The output features from the dynamic convolution layer are utilized for distance and loss computation.
To be consistent with the size of dynamic kernel which is $512\times512\times1\times1$, the output size of controller has been set to $2048\times16\times8/1024\times16\times16$ for person/vehicle instances.

We argue that the self-guided dynamic convolution module is favorable to capture the instance-level discriminative feature.
As it can be found in Figure~\ref{fig:heatmap}, compared with the holistic-level features from global branch, features produced by self-guided dynamic convolution tend to capture discriminative parts which are crucial for distinguishing a particular instance from others. 

\begin{figure}[t]
	\begin{center}
		\includegraphics[width=1.0\linewidth]{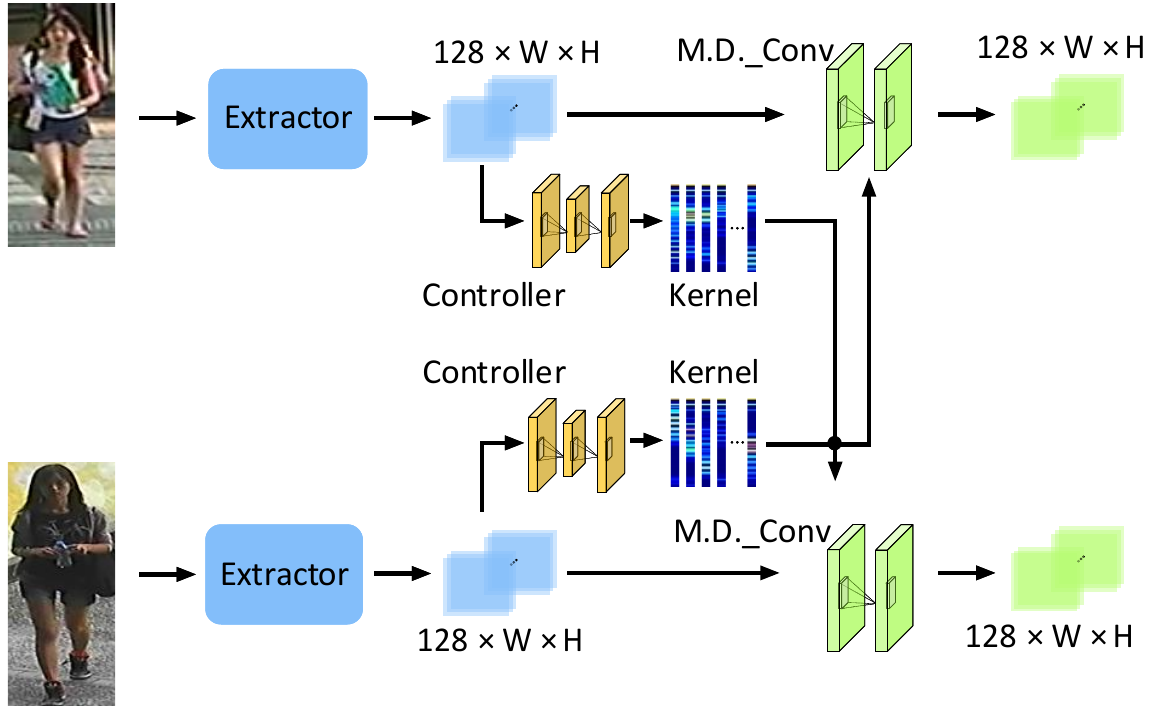}
	\end{center}
	\vspace{0.05cm}
	\caption{
		The structure diagram of our Mutual-Guided Dynamic Convolution Module. The ``M.D.\_Conv" represents Mutual-Guided Dynamic Convolution layer. In our experiments, the dimension of the input/output feature of Mutual-Guided Dynamic Convolution layer is designed as 128 (256) for person (vehicle) instance. The ``Extractor" and ``Controller" shown above shared parameters.}
	\label{fig:distance_matrix}
\end{figure}

\subsection{Mutual-guided Dynamic Branch}
In this subsection, the proposed mutual-guided dynamic branch will be illustrated in detail.  
Firstly, we would like to elaborate on the motivation of constructing this branch. 
In order to capture the subtle but mutually discriminative cues between input images, we not only ought to depend on the input image itself but should turn to the image to be compared as well.   
To this end, during the feature extraction phase we regard each pair of input images as independent unit in which they mutually provide dynamic convolution kernels to extract pair-aware instance-specific features.


\begin{figure}[t]
	\begin{center}
		\includegraphics[width=1.0\linewidth]{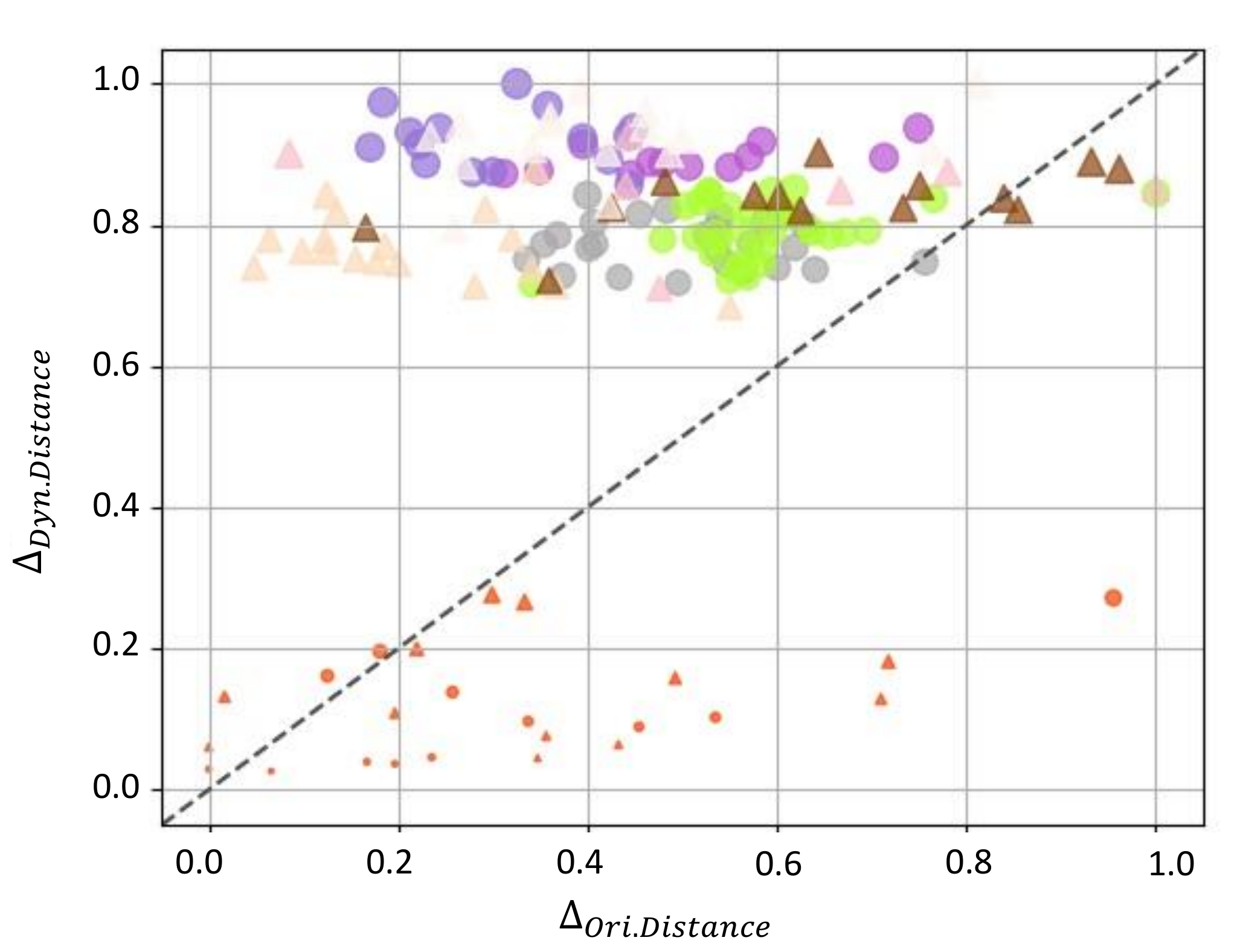}
	\end{center}
	\vspace{-5mm}
	\caption{
	    We visualize the similarity matching results belonging to single query image.
	    {\bf{Different colors represent specific identity in which the red ones denote positive instances.}}
	    The axis x/y represents the distances of feature before/after the dynamic convolution module.
	    Distances shown above have been normalized.
	    The circles/triangles represent features generated from mutual-guided/self-guided dynamic branch.
	    Beyond that, the sizes of circles/triangles are proportional to the absolute feature distance after dynamic module.
        It can be clearly found that both the Self-Guided Dynamic Convolution Module and the Mutual-Guided Dynamic Convolution Module assist to reduce the distance between positive pairs and vastly increase that between negative instances.
	}
	\label{fig:mutual}
\end{figure}

In addition to the motivation of constructing this branch, the structure of mutual-guided dynamic convolution module we adopted in this branch will be demonstrated in detail.
The sketch of our mutual-guided dynamic convolution branch is shown in Figure~\ref{fig:distance_matrix}.
Similar as the self-guided dynamic branch, we firstly apply a channel compression module in this branch.
The output feature dimension of the channel compression module has been set as $128/256$ for person/vehicle instance. 
Beyond the channel compression module, a $1\times1$ dynamic convolution layer whose input and output dimensions are both $128/256$ has been employed in this branch. 
Whilst, the controller module shown in Figure~\ref{fig:distance_matrix} is also a shallow network which has the same structure as described above.
To be consistent with the size of dynamic kernel, the output size of controller is set to $128\times16\times8$ for person instances while for vehicle instances the output size is set to $256\times16\times16$.


Meanwhile, the training scheme adopted in our mutual-guided dynamic branch would be described. 
Primarily, we would like to explain the reason we abandon the ID Loss in this branch.
Because of the pair-aware property in this branch, each input image will be represented $N$ times via the different kernels produced by all $N$ individuals in the mini-batch. In this case, $N\times N$ ID losses need to be calculated which can lead to a relatively huge computational consumption. Moreover, the main purpose of mutual-guided branch is to cross-refer image pairs and exploit relevant semantic areas, which means that the obtained features could be suitable for matching procedure but less salient for single ID identification.
Therefore, it is clear that adopting ID Loss in this branch is uneconomical and meaningless.
Fortunately, the metric learning methods such as Triplet Loss definitely fit our mutual-guided dynamic mechanism.
The essence of Triplet Loss can be regarded as measuring the similarity between pair of images which denotes it can be directly adopted in this branch.

As it can be found in Figure~\ref{fig:distance_matrix}, pairwise input images mutually provide dynamic kernels for further feature refining during feature extracting phase. The distance between feature $f_i$ and $f_j$ can be denoted as: 
\begin{equation}
d_{i,j}=D((K_{i}f_{j}),(K_{j}f_{i}))
\end{equation}\label{fun:distance}
\noindent where the $D()$ denotes the Euclidean distance which we applied in our algorithm for similarity metric. In our algorithm, the same distance computation scheme is utilized in the test phase, as well.

To prove our dynamic module facilitate identifying, We visualize the similarity matching results belonging to single query image in Figure~\ref{fig:mutual}. 
It is obvious that dynamic modules assist to reduce the distance between query image and gallery belonging to the same identity and vastly increase the distance between different instances.
It denotes that our proposed dynamic branches are effective measures to distinguish specific identity from others.

\begin{table}[t]
\centering
\setlength{\belowcaptionskip}{-0.cm}
\resizebox{0.5\textwidth}{!}{
\begin{tabular}{l|c c c c}
			\hline
			Methods & Dataset & mAP & CMC-1 & CMC-5\\
			\hline
			Baseline & Market1501 & $64.6$ & $81.8$ & $90.2$ \\
			Baseline & DukeMTMC-reID & $69.2$ & $83.3$ & N/A \\
			Baseline & CUHK03(D.) & $69.3$ & $84.4$ & $92.2$ \\
			Baseline & CUHK03(L.) & $71.0$ & $84.4$ & N/A \\
			Baseline & VeRi & $71.8$ & $84.9$ & N/A \\
			Baseline & VehicleID & $78.4$ & $88.7$ & N/A \\
			\hline
			IPAD & Market1501 & $65.5$ & $82.6$ & N/A \\	
			IPAD & DukeMTMC-reID & $72.9$ & $85.8$ & N/A \\
			IPAD & CUHK03(D.) & $73.1$ & $86.5$ & $93.1$ \\
			IPAD & CUHK03(L.) & $73.4$ & $87.1$ & N/A \\
			IPAD & VeRi & $73.7$ & $87.7$ & N/A \\
			IPAD & VehicleID & $74.3$ & $87.7$ & N/A \\
			\hline
		\end{tabular}
		}

\caption{The experimental results of  ``Baseline" and ``IPAD" on diverse mainstream ReID datasets. It can be found that our IPAD sightly promote the capability of re-identification module in all these ReID datasets.}
\label{tab:duke}
\end{table}

\begin{table}[t]
\centering
\setlength{\belowcaptionskip}{-0.cm}
\resizebox{0.5\textwidth}{!}{
\begin{tabular}{l|c c c c}
			\hline
			Methods & Backbone & mAP & CMC-1 & CMC-5\\
			\hline
			Baseline & VGG19\_bn & $64.6$ & $81.8$ & $90.2$ \\
			Baseline & Resnet34 & $69.2$ & $83.3$ & N/A \\
			Baseline & Resnet50 & $69.3$ & $84.4$ & $92.2$ \\
			Baseline & SE-Resnet50 & $71.0$ & $84.4$ & N/A \\
			Baseline & MobileNet\_V2 & $71.8$ & $84.9$ & N/A \\
			\hline
			IPAD & VGG19\_bn & $65.5$ & $82.6$ & N/A \\	
			IPAD & Resnet34 & $72.9$ & $85.8$ & N/A \\
			IPAD & Resnet50 & $73.1$ & $86.5$ & $93.1$ \\
			IPAD & SE-Resnet50 & $73.4$ & $87.1$ & N/A \\
			IPAD & MobileNet\_V2 & $73.7$ & $87.7$ & N/A \\
			\hline
		\end{tabular}
		}

\caption{The performance comparison on DukeMTMC-reID dataset. In order to verify the performance gain of our ``IPAD" is consistent with various backbone, we exchange several mainstream backbone model to the Baseline Module and our IPAD. It can be found that our ``IPAD" can always achieve performance improvement while combining with all  these mainstream backbone model.It denotes that our ``IPAD" is pluggable and compatible with various backbone models.}
	\label{tab:duke}
\end{table}

\begin{table}[t]
\centering
\setlength{\belowcaptionskip}{-0.cm}
\resizebox{0.45\textwidth}{!}{
\begin{tabular}{l|c c c c}
			\hline
			Methods & mAP & CMC-1 & CMC-5 & CMC-10\\
			\hline
			Baseline & $64.6$ & $81.8$ & $90.2$ & $96.1$ \\
            IPAD\_V1 & $77.4$ & $87.2$ & $94.6$ & $96.1$ \\	
            IPAD\_V2 & $78.3$ & $88.6$ & $94.6$ & $96.1$ \\	
            IPAD & $78.3$ & $88.6$ & $94.6$ & $96.1$ \\	
			\hline
		\end{tabular}
		}

\caption{The ``IPAD\_V1" is trained with the Global Branch and the Mutual-Guided Dynamic Branch. The ``IPAD\_V2" represents the model trained with Global Branch and the Self-Guided Dynamic Branch. In order to show how each branch of the proposed model contributes to the final performance, we gradually add components during training phase based on the baseline naive global branch and compare their performance on DukeMTMC-reID dataset.}
\label{tab:duke}
\end{table}

\subsection{Discussion}
{\bf{Comparison with attention-based methods.}}

To be best of our knowledge, quit a few existing attention-based algorithms have been constructed with similar motivation as our IPAD.
Most of these methods seek to capture discriminative representations via weighting salient local human part.
Despite the advances of attention-based modules, we argue that they still face sightly deficiencies. 
Existing attention-based methods have solved "where to focus" which have been proved by visualization and experimental results.
However, they neglect exploring more effective feature augment function rather than directly weigthing input convolutional feature with generated attention maps. 

To this end, in this paper we devote to explore a robust visual feature augment function which is capable to convert the vanilla convolutional feature into much more effective representations. 
The dynamic convolution mechanism employed in our framework can flexibly provide effective feature augment function and it is well compatible with ReID module, thus we construct multi-guidance dynamic convolution branch to promote capability of our module.
In order to verify the superiority of our proposed IPAD over attention-based methods, we carefully design a series of  ablation studies. The experiments detail will be illustrated in section~\ref{sec_abla}.



\section{Experiments}\label{sec_expr}

In this section, we conduct comprehensive experiments to verify the effectiveness of our proposed algorithm. 
In Section~\ref{sec_detail} we elaborate on the evaluation protocols and implementation details. In Section~\ref{sec_dataset}, the ReID datasets we evaluate our algorithm on will be introduced roughly. Besides, we illustrate ablation experiments in detail in Section~\ref{sec_abla}. Meanwhile, our proposed algorithm will be compared with other state-of-the-art algorithms in section~\ref{sec_comp}. 
To clarify, firstly we give the evaluation protocols and
implementation details.

\subsection{Evaluation Protocols and Implementation Details}\label{sec_detail}
In our experiments, we adopt Resnet50 pre-trained on ImageNet as backbone. In the training phase, we flip images horizontally with a probability of $0.5$ for data augmentation. 
After that, we resize each person (vehicle) image into 256 × 128 (256) pixels and pad the resized image 10 pixels with zero values. Then we randomly crop it into a 256 × 128 (256) rectangular image.
The mini-batch size is set to 16 × 4 = 64, with 16 identities and 4
images for each ID. We adopt Adam optimizer whose initial learning rate is $3.5\times10^{-4}$ and is decreased by 0.1 at the 40th epoch and 70th epoch respectively. In our experiments, all models are trained for 160 epochs. During the test phase, we both extract the features of original images and the horizontally flipped versions, then use the average of these as the final features. 
After that, we fuse the distance matrices from three branches of IPAD for the final matching. 
We adopt the popular mean Average Precision
(mAP) and Cumulative Matching Cure (CMC) as evaluation criteria.

\subsection{Datasets}\label{sec_dataset}
In this subsection we introduce the widely used ReID datasets being employed in our experiments.

\noindent{\bf{Person ReID Datasets:}}

\noindent{\bf{Market1501~\cite{zheng2015scalable}}} is constructed with $32,668$ images belonging to $1,501$ instances, among which $12,936$ images are separated into training set, $3,368$ images are in the query set and $15,913$ images are in the gallery set.
{\bf{DukeMTMC-reID~\cite{ristani2016performance}}} is also a mainstream dataset for person ReID, which consists of $36,411$ images belonging to $1,812$ instances in which $16,522$ images are in the training set, $2,228$ images are in query set, and $17,661$ images are in gallery set. {\bf{CUHK03~\cite{li2014deepreid}}} is composed of two subset one of them labeled pedestrian boxes manually (CUHK03-Label), the other one is constructed with the assistance of detected algorithm (CUHK03-Detect). CUHK03-Label/CUHK03-Detect consists of $7,368$/$7,365$ training images, $1,400$/$1,400$ query images and $5,328$/$5,332$ gallery images.

\noindent{\bf{Vehicle ReID Datasets:}}

\noindent{\bf{VehicleID~\cite{liu2016relative}}} is a large scale vehicle ReID dataset. It consists of $221,763$ images belonging to $26,267$ vehicle instances. It has defined three test sets according to their size (i.e. small, medium and large). 
{\bf{VeRi776~\cite{liu2016deep}}} contains $37,778$ images in training set, $1,678$ images in query set and $9,901$ images in gallery set.

\begin{figure*}[t]
	\begin{center}
		\includegraphics[width=1.0\linewidth]{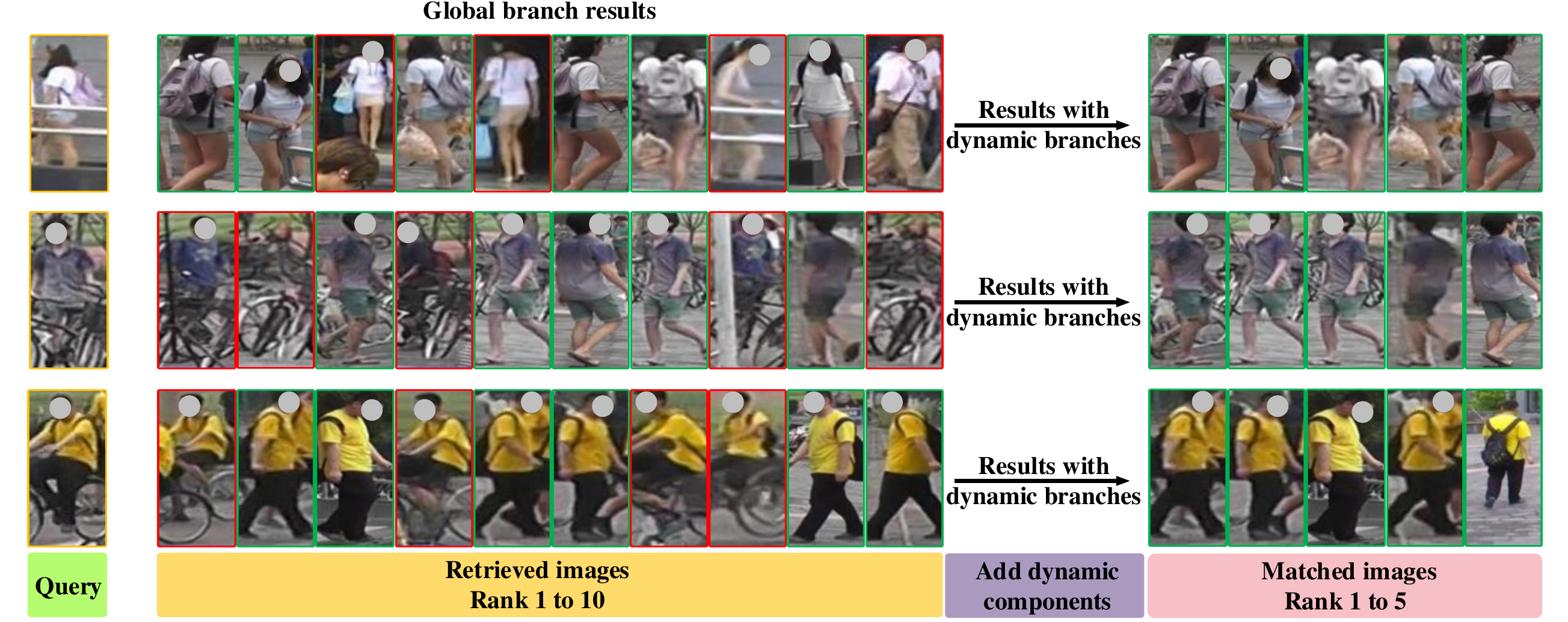}
	\end{center}
	\vspace{0mm}
	\caption{
		Three examples of how dynamic branches assist to enhance model capability. The chosen examples are difficult for identifying and return many falsely matched images. Green box denotes same ID as query while red box represents different ID from query.}
	\label{fig:failure_case}
\end{figure*}

\subsection{Ablation Experiments}\label{sec_abla}
In this subsection, we design several experiments to fully test our proposed model from various aspects. 

{\bf{Comparisons with baseline model.}} The first group of Table~\ref{tab:ablation} shows comparisons of our model with baseline model in several widely used ReID datasets. In our experiments, we treat the vanilla Resnet50 trained with Triplet Loss and ID Loss as the  ``Baseline" model. As it can be found in Table~\ref{tab:ablation}, our Instance and Pair-Aware Dynamic Network (IPAD) significantly improves model performance both in the vehicle and person ReID dataset \ie  $3.8$/$2.7$ mAP/CMC-1 on average. 
It denotes the proposed IPAD is capable and effective to identify vehicle and person instance.

{\bf{Consistency in performance gain with various backbones.}}
To verify the performance gain of our IPAD is consistent with various backbone, we exchange several mainstream backbone models including VGG19\_bn, Resnet34, Resnet50, SE-Resnet50 and MobileNetv2 to IPAD and the baseline model. After that, we train and test the exchanged versions in DukeMTMC-reID dataset. The experimental results are shown in the second group of  Table~\ref{tab:ablation}. It can be found that our IPAD can always achieve performance improvement while combining with all these mainstream backbone model. It denotes that our IPAD is pluggable and compatible with various backbone models.

{\bf{Effects  of  branches.}} To show how each branch of the proposed model contributes to the final ReID performance, we gradually add components during training phase based on the baseline naive global branch and compare the performance improvements on DukeMTMC-reID dataset.
The experiential results are shown in the third group of Table~\ref{tab:ablation}. 
Compared with the baseline model, ``IPAD\_V1" is trained with an additional mutual-guided dynamic branch. It shows that the mutual-guided dynamic branch can bring an additional improvement of $1.0$\% mAP and $0.5$\% CMC-1 to the baseline module.
Meanwhile, the ``IPAD\_V2" model is trained with an additional self-guided dynamic branch.
As it can be found in Table~\ref{tab:ablation}, ``IPAD\_V2" model achieves a performance improvement of $1.9$\% mAP, and $1.9$\% CMC-1.
In the integrated IPAD model, all the three branches are combined together. The results of IPAD model are significantly better than those of other compared models.
To be exact, it can get an additional improvement of $3.1$\% mAP, and $2.0$\% CMC-1 over the baseline method.
The experimental results denote that both the mutual-guided dynamic branch and the self-guided dynamic benefit instance identifying while they can also mutually provide complementary information. 

Beyond the experimental results, we show several matched results of the global branch and the integrated IPAD in Figure~\ref{fig:failure_case}. As it shown in Figure~\ref{fig:failure_case}, global module attaches more importance to the holistic-level similarity which leads to the shown falsely retrieved cases. For the matched results of our integrated model, it sightly surpasses the naive global branch via the effective and instance-specific features extracted by employed dynamic branches.


\begin{table*}[t]
\resizebox{1.0\textwidth}{!}{
\begin{tabular}{p{3cm} | c |c c c| c c| c c| c c }
			\hline
			 & & \multicolumn{3}{c|}{VeRi} &\multicolumn{6}{c}{VehicleID}\\
			 & & \multicolumn{3}{c|}{} &
			 \multicolumn{2}{c}{Small} & \multicolumn{2}{c}{Medium} &
			 \multicolumn{2}{c}{Large} \\
			\hline
			method & Reference & mAP & CMC-1 & CMC-5 & mAP & CMC-1 & mAP & CMC-1 & mAP & CMC-1 \\
			\hline
			NuFACT~\cite{liu2018provid} & TMM2018 & $50.9$ & $81.1$ & N/A & N/A & $48.9$ & N/A & $43.6$ & N/A & $38.6$ \\
			C2F~\cite{guo2018learning} & AAAI2018 & N/A & N/A & N/A & $63.5$ & $61.1$ & $60.0$ & $56.2$ & $53.0$ & $51.4$ \\
			SDC-CNN~\cite{zhu2018shortly} & ICPR2018 & $53.5$ & $83.5$ & $92.6$ & $63.5$ & $57.0$ & $57.1$ & $50.6$ & $49.7$ & $42.9$ \\
			GSTE~\cite{bai2018group} & TMM2018 & $59.5$ & $96.2$ & {\bf{99.0}} & $75.4$ & $75.9$ & $74.3$ & $74.8$ & $72.4$ & $74.0$ \\
			VAMI~\cite{zhou2018viewpoint} & CVPR2018 & $61.3$ & $85.9$ & $91.8$ & N/A & $63.1$ & N/A & $52.9$ & N/A & $47.3$ \\
			PMSM~\cite{sun2018part} & ICPR2018 & N/A & N/A & N/A & $64.2$ & N/A & $57.2$ & N/A & $51.8$ & N/A \\ 
			RAM~\cite{liu2018ram} & ICME2018 & $61.5$ & $88.6$ & $94.0$ & N/A & $75.2$ & N/A & $72.3$ & N/A & $67.7$ \\
			\hline
			FDA-Net~\cite{lou2019veri} & CVPR2019 & $55.5$ & $84.3$ & $92.4$ & N/A & N/A & $65.3$ & $59.8$ & $61.8$ & $55.5$ \\
			EALN~\cite{liu2019embedding} & TIP2019 & $57.4$ & $84.4$ & $94.1$ & $77.5$ & $75.1$ & $74.2$ & $71.8$ & $71.0$ & $69.3$ \\
			MTML~\cite{rajamanoharan2019multi} & CVPR2019 & $64.6$ & $92.3$ & $95.7$ & N/A & N/A & N/A & N/A & N/A & N/A \\
			MoV1+BS~\cite{Kumar2019vehicle} & IJCNN2019 & $67.6$ & $90.2$ & $96.4$ & \textit{84.2} & $78.8$ & \textit{79.1} & $73.4$ & \textit{75.5} & \textit{69.3} \\
			\hline
			PVEN~\cite{meng2020parsing} & CVPR2020 & \textit{79.5} & \textit{95.6} & \textit{98.4} & N/A & \bf{84.7} & N/A & \bf{80.6} & N/A & \bf{77.8} \\
			\hline
			IPAD & $-$ & \bf{82.4} & \bf{96.6} & \bf{99.0} & \bf{89.6} & \textit{84.6} & \bf{86.1} & \textit{80.4} & \bf{83.6} & \bf{77.8} \\
			\hline

		\end{tabular}
		}
\caption{Performance comparison of our IPAD with other state-of-the-art algorithms on VeRi776 and VehicleID datasets. It can be found that our IPAD outperforms other compared algorithms on VeRi776. Meanwhile, our IPAD achieved a comparable performance on VehicleID due to the limited image view-angle changes in VehicleID.}
	\label{tab:vehicle}
\end{table*}

\begin{table}[t]
\centering
\setlength{\belowcaptionskip}{-0.0cm}
\resizebox{0.5\textwidth}{!}{
\begin{tabular}{l|c c c c}
			\hline
			Methods & Reference & mAP & CMC-1 & CMC-5\\
			\hline
			DuATM\cite{si2018dual} & CVPR2018 & $64.6$ & $81.8$ & $90.2$ \\
			PCB+RPP~\cite{sun2018beyond} & ECCV2018 & $69.2$ & $83.3$ & N/A \\
			PABR~\cite{suh2018part} & ECCV2018 & $69.3$ & $84.4$ & $92.2$ \\
			SPReID~\cite{kalayeh2018human} & CVPR2018 & $71.0$ & $84.4$ & N/A \\
			Mancs~\cite{wang2018mancs} & ECCV2018 & $71.8$ & $84.9$ & N/A \\
			MGN~\cite{wang2018learning} & MM2018 & $78.4$ & $88.7$ & N/A \\
			\hline
			PGFA~\cite{miao2019pose} & ICCV2019 & $65.5$ & $82.6$ & N/A \\	
			CAMA~\cite{yang2019towards} & CVPR2019 & $72.9$ & $85.8$ & N/A \\
			P$^{2}$ -Net~\cite{guo2019beyond} & ICCV2019 & $73.1$ & $86.5$ & $93.1$ \\
			IANet~\cite{hou2019interaction} & CVPR2019 & $73.4$ & $87.1$ & N/A \\
			CASN+PCB~\cite{zheng2019re} & CVPR2019 & $73.7$ & $87.7$ & N/A \\
			AANet~\cite{tay2019aanet} & CVPR2019 & $74.3$ & $87.7$ & N/A \\
			DSA-reID~\cite{zhan2019densely} & CVPR2019 & $74.3$ & $86.2$ & N/A \\
			\hline
			SCSN~\cite{chen2020salience} & CVPR2020 & $79.0$ & {\bf91.0} & N/A \\
			\hline
			IPAD & - & \bf{79.5} & $88.7$ & \bf{95.0} \\
			\hline
		\end{tabular}
		}

\caption{The performance comparison on DukeMTMC-reID dataset. It can be found that our IPAD It can be found that our IPAD achieves promising performance.}
	\label{tab:duke}
\end{table}

\begin{table}[t]
\centering
\resizebox{0.5\textwidth}{!}{
\begin{tabular}{l|c c c c}
			\hline
			Methods & Reference & mAP & CMC-1 & CMC-5\\
			\hline
			DuATM\cite{si2018dual} & CVPR2018 & $76.6$ & $91.4$ & $97.1$ \\
			PCB+RPP~\cite{sun2018beyond} & ECCV2018 & $81.6$ & $93.8$ & $97.5$ \\
			PABR~\cite{suh2018part} & ECCV2018 & $79.6$ & $91.7$ & $96.9$ \\
			SPReID~\cite{kalayeh2018human} & CVPR2018 & $81.3$ & $92.5$ & N/A \\
			Mancs~\cite{wang2018mancs} & ECCV2018 & $82.3$ & $93.1$ & N/A \\
			MGN~\cite{wang2018learning} & MM2018 & $86.9$ & $95.7$ & N/A \\
			\hline
			PGFA~\cite{miao2019pose} & ICCV2019 & $76.8$ & $91.2$ & N/A \\	
			CAMA~\cite{yang2019towards} & CVPR2019 & $84.5$ & $94.7$ & $98.1$ \\
			P$^{2}$ -Net~\cite{guo2019beyond} & ICCV2019 & $85.6$ & $95.2$ & $98.2$ \\
			IANet~\cite{hou2019interaction} & CVPR2019 & $83.1$ & $94.4$ & N/A \\
			CASN+PCB~\cite{zheng2019re} & CVPR2019 & $82.8$ & $94.4$ & N/A \\
			AANet~\cite{tay2019aanet} & CVPR2019 & $83.4$ & $93.9$ & N/A \\
			DSA-reID~\cite{zhan2019densely} & CVPR2019 & $87.6$ & $95.7$ & N/A \\
			\hline
			SCSN~\cite{chen2020salience} & CVPR2020 & $88.5$ & $95.7$ & N/A \\
			RGA-SC~\cite{zhang2020relation} & CVPR2020 & $88.4$ & {\bf96.1} & N/A \\
			\hline
			IPAD & - & \bf{88.7} & $95.2$ & \bf{99.1} \\
			\hline
		\end{tabular}
		}
\caption{The performance comparison on Market-1501 dataset. It can be found that our IPAD achieves promising performance.}
	\label{tab:market}
\end{table}

{\bf{Comparison with attention-based methods.}}

In order to verify the superiority of our proposed IPAD over those attention-based methods, we exchange the self-guided dynamic convolution layer and the mutual-guided dynamic convolution layer in our IPAD to self-attention and co-attention mechanism. 
To compare in a fair setting, we adopt the same implementation details to train and test our IPAD and the attention-based version on DukeMTMC-reID. 
Besides, the feature dimension of self-attention branch (co-attention branch) has also been compressed to 512 (128) to be consistent with our dynamic branches.
The experimental results show that the mAP/CMC-1 of our IPAD is $0.8/0.7$ higher than it of the attention-based version which proves our argument.

\subsection{Comparison with State-of-the-art Approaches}\label{sec_comp}
In this subsection, we compare the proposed IPAD with other state-of-the-art algorithms in several mainstream datasets to verify the effectiveness of our proposed model.

{\bf{Evaluation on the vehicle ReID datasets}}
In Table~\ref{tab:vehicle}, we show the experimental results of our IPAD and other compared methods on VeRi776 and VehicleID datasets. In these compared algorithms, several algorithms require additional annotation information such as MTML and RAM. The other some are based on the view-angle alignment just like PVEN and VAMI. Beyond that, the MOV1+BS is a strong ReID baseline model and the HPGN aims to explore the spatial significance of feature maps.

All compared algorithms aim at mining discriminative cues only relying on the input images itself while our IPAD novelly proposes a mutual-guided dynamic branch to extract instance-specific features through pair-aware mutual-referring between input images.
Therefore, our IPAD achieved sightly better performance than these compared algorithms on VeRi776 dataset. 
However, different from the VeRi776, the image view-angles in VehicleID are very limited whose distribution concentrates in the front and behind views.
In addition, the appearance gap between different views of vehicle instances is so huge that the crucial cross-referring mechanism in the mutual-guided branch can hardly work to its best. 
Nevertheless, the performance of our IPAD
on VehicleID is still comparable with those algorithms.

{\bf{Evaluation on the person ReID dataset}} In Table~\ref{tab:duke} and Table~\ref{tab:market}, we show the comparison of our IPAD with other compared algorithms on DukeMCMT-reID and Market1501. Some of these compared algorithms focus on extracting effective feature through attention-based methods such as RGA-SC and CAMA. Besides, several algorithms aim at enhancing algorithms capability by using the alignment-based methods just like PABR. Meanwhile, some other algorithms such as AANet and DSA-reID hammer at generating more discriminative feature representations with the assistance of additional annotation information. 

For our IPAD, we do not utilize any additional annotation while we both apply a self-guided dynamic branch to mining effective feature and construct a mutual-guided dynamic branch to generate pair-aware instance-specific features. It can be found that the mAP and CMC-5 of our IPAD outperform these compared methods while the CMC-1 of our IPAD is comparable with others.

\section{Conclusion}\label{sec_conc}
In this paper, we propose an Instance and Pair-Aware Dynamic Network model to extract expressive features via customized filters for each input images and capture the subtle but pairwise mutually discriminative visual cues.
The proposed framework is able to generate instance-specific features for each input instance by the self-guided dynamic competent while it can also capture pair-aware discriminative features through employing the mutual-guided dynamic module.
Extensive visualisation and experimental results show that the IPAD can promote the capability of ReID module while achieve promising results on many mainstream ReID benchmark.
In some datasets our algorithm outperforms state-of-the-art methods and in the other employed datasets our algorithm achieves a comparable performance.


%

\ifCLASSOPTIONcaptionsoff
  \newpage
\fi



%

{\small
	\bibliographystyle{ieee}
	\bibliography{main}
}

\end{document}